\documentclass[runningheads]{llncs}

\usepackage{graphicx}%
\usepackage{multirow}%
\usepackage{amsmath,amssymb,amsfonts}%
\usepackage{mathrsfs}%
\usepackage{manyfoot}
\usepackage{booktabs}
\usepackage{algorithm}
\usepackage{algorithmic}
\usepackage{listings}

\usepackage{marvosym} 
\usepackage{enumitem} 
\usepackage[inkscapelatex=false]{svg}
\usepackage{hyperref}
\hypersetup{
    colorlinks=true,   
    linkcolor=blue,    
    citecolor=blue,    
    urlcolor=blue      
}

\usepackage{subfig} 
\newcommand{\vnudge}{\vspace*{-.2in}} 

\begin{document}
%
\title{
GI-SMN: Gradient Inversion Attack against Federated Learning without Prior Knowledge}
\titlerunning{Gradient Inversion Without Prior Knowledge}

\author{Jin Qian\inst{1,2}\and Kaimin Wei\inst{1,2}\and Yongdong Wu\inst{1,2}\and Jilian Zhang\inst{1,2}\and Jipeng Chen\inst{3}\and Huan Bao\inst{1,2}}


\authorrunning{J.Qian et al.}
%
\institute{College of Information and Technology, Jinan University, Guangzhou, 510632, China \email{cswei@jnu.edu.cn} \and Guangdong Key Laboratory of Data Security and Privacy Protection, Guangzhou, 510632, China \and School of Computer Science, Beijing University of Posts and Telecommunications, Beijing, 100876, China}
%
\maketitle              

\begin{abstract}
Federated learning (FL) has emerged as a privacy-preserving machine learning approach where multiple parties share gradient information rather than original user data. Recent work has demonstrated that gradient inversion attacks can exploit the gradients of FL to recreate the original user data, posing significant privacy risks. However, these attacks make strong assumptions about the attacker, such as altering the model structure or parameters, gaining batch normalization statistics, or acquiring prior knowledge of the original training set, etc. Consequently, these attacks are not possible in real-world scenarios. To end it, we propose a novel Gradient Inversion attack based on Style Migration Network (GI-SMN), which breaks through the strong assumptions made by previous gradient inversion attacks. The optimization space is reduced by the refinement of the latent code and the use of regular terms to facilitate gradient matching. GI-SMN enables the reconstruction of user data with high similarity in batches. Experimental results have demonstrated that GI-SMN outperforms state-of-the-art gradient inversion attacks in both visual effect and similarity metrics. Additionally, it also can overcome gradient pruning and differential privacy defenses.


\end{abstract}

\section{Introduction}
As deep learning technology advances, an increasing number of deep learning applications depend on vast quantities of high-quality data for effective model training. However, due to the presence of data silos among different entities \cite{li2022federated}, individual entities often have access to only limited or low-quality data, which complicates the training of deep learning models. Although pooling data from all entities could address this issue, it raises concerns over issues like copyright infringement and privacy breaches \cite{li2022federated}. In response, FL is proposed, which avoids centralizing all data in one place for model training \cite{li2022federated}. FL employs a parameter server to send a global model to each user, who then partially trains it on their local device before transmitting updated gradient information back to the central parameter server for aggregation, thereby training the global model. This decentralized training approach prevents the transmission of personal data to centralized servers or other users, ensuring data privacy and security.

\begin{figure}[t]
    \centering
    \includegraphics[width=0.7\columnwidth]{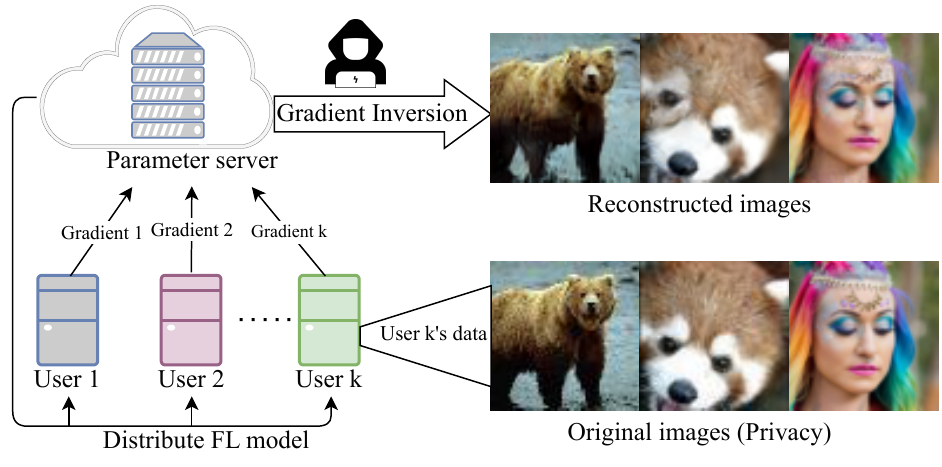}
    \caption{An example of gradient inversion attack. The attacker reconstructs the original user data by utilizing the obtained gradient information.
    }\label{fig:main} \vnudge
\end{figure}

Recent research has demonstrated that merely exchanging gradients does not sufficiently protect compromised user privacy in federated learning \cite{zhu2019deep}. In gradient inversion attacks, attackers can recover labels \cite{geng2023improved}, attributes \cite{wang2023privacy}, and even clear images of the original data by exploiting the shared gradient information. As displayed in Fig. \ref{fig:main}, the attacker reconstructs the FL participant's original training images from the obtained gradient and FL model. Although participants only upload the gradient information of their training data, it still leads to privacy breaches \cite{wang2023privacy}. The gradient inversion attack is founded on a straightforward principle: the closer the gradients of two data points are, the more similar those data points become \cite{yang2022using}. For instance, during image reconstruction, a federated learning participant possesses a private and original image, and the attacker aims to reconstruct the specific original image. The attacker begins by generating a dummy image and then minimizes the distance between the corresponding gradients of the two images. This process enables the creation of a dummy image that is highly similar to the original image, thereby effectively reconstructing the private user data.

Despite recent advancements in gradient inversion attacks against FL, there are still notable deficiencies, such as low similarity of reconstructed data \cite{yin2021see}, inability to reconstruct data in batches \cite{wen2022fishing}, etc. In particular, some gradient inversion attacks assume that attackers can modify the weights or structures of federated learning models \cite{fowl2021robbing,wen2022fishing}. An attacker with such superpowers would be easily detected during an actual attack \cite{chen2020training}. Furthermore, some gradient inversion attacks rely heavily on additional prior knowledge to aid the reconstruction of user data. For example, they utilize the batch normalized statistics to recreate images, employ the selected prior images as the optimal starting point for reconstructing images \cite{hatamizadeh2023gradient}, and use the generative models trained on FL users' local data \cite{jeon2021gradient}. However, it is difficult to find these super-powered attackers and acquire the idealized prior knowledge in FL. The training data is usually stored locally, making it difficult for attackers to obtain prior images or pre-trained models without access. Meanwhile, the batch normalization statistics are not uploaded to the parameter server, making them unavailable to attackers \cite{huang2021evaluating}. Therefore, it is necessary to investigate gradient inversion attacks that require neither a powerful attacker nor idealized prior knowledge to enhance the privacy-preserving functionalities of federated learning.

In this paper, we propose a novel Gradient Inversion attack based on Style Migration Network, which requires neither super-attackers nor idealized prior knowledge. In GI-SMN, the latent encoding, the input of the style migration network, is carefully optimized to significantly reduce the required parameter optimization space, thereby avoiding the need for additional prior knowledge. Moreover, a series of regularization terms are integrated into the loss function to assist gradient matching, thereby reconstructing target data with high similarity. Extensive experiments have been conducted, and results have demonstrated that GI-SMN outperforms state-of-the-art gradient inversion attack methods regarding visual effect and similarity metrics. Additionally, it also can overcome gradient pruning and differential privacy defenses. The main contributions of this paper are as follows:
\begin{enumerate}[label=(\arabic*)]
    \item Unlike traditional gradient inversion attacks, GI-SMN overcomes the dependence on powerful attackers and idealized prior knowledge, making the attack more threatening.
        
    \item GI-SMN can recreate original data with high similarity in batches using a style migration network and a series of regularization terms.
    
    \item GI-SMN surpasses state-of-the-art gradient inversion attacks in visual effect and similarity metrics.
    
    \item This paper demonstrates that gradient pruning and differential privacy are not effective defenses against privacy breaches in federated learning.
\end{enumerate}

\section{Related Work}\label{sec:related}
This section provides an overview of recent gradient inversion attacks, which can be separated into two categories based on how to reconstruct images: iterative optimization-based and analytical reasoning-based.

\subsection{Gradient Inversion Attacks Based On Iterative Optimization}
In iterative optimization-based gradient inversion attacks, attackers create dummy data and feed it into the FL model to produce dummy gradients. The dummy data are continuously optimized by calculating the distance between the true and dummy gradients until the private data is reconstructed. Moreover, dummy data initialization is crucial for data reconstruction, and poor initialization will lead to a failed data reconstruction \cite{geiping2020inverting}. The dummy data is initialized by a random Gaussian \cite{geiping2020inverting,yin2021see,zhu2019deep}, constant \cite{wang2020sapag}, or uniform distribution \cite{jin2021cafe} before iterative optimization. During the repeated optimization procedure to extract the final gradients of dummy data, attackers supposedly have easy access to the true gradient and FL model. However, various gradient inversion attacks have varying assumptions about attackers' capabilities \cite{jeon2021gradient}. Gradient matching is a research topic in iterative optimization, which aims to minimize the difference between gradients by optimizing metrics and regularization terms to recover more realistic data \cite{zhang2022survey}. Loss functions employing Euclidean distance \cite{jin2021cafe} and cosine \cite{geiping2020inverting,huang2023neurogenesis} to calculate the separation between the dummy and true gradients. In addition, regularization items can be split into two groups: improving fidelity and correcting image position. Prior information, such as BN statistics, can significantly enhance the quality of data reconstruction \cite{wang2019beyond}, but is difficult to obtain in reality. Therefore, investigating attacks that do not require additional prior knowledge and ensure stability in gradient inversion is essential.

\subsection{Gradient Inversion Attacks Based On Analytical Inference}
Instead of iterative optimization-based attacks, analytical inference-based gradient inversion attacks reconstruct data from gradient information by parsing formulas \cite{fowl2021robbing}. These attacks typically deal with limited network types. However, they cannot handle ResNet's pooling layer and residual links \cite{zhang2022survey}, resulting in cumulative errors and reconstructed data that are distant from the originals. Since analytical inference-based attacks rely on the completeness of the gradient information, they will fail to reconstruct data when perturbations are added or insufficient gradient information is employed. In addition, since it is challenging for analytical inference-based attacks to reconstruct data just from gradients, they often require additional prior knowledge. Fowl et al. \cite{fowl2021robbing} incorporated many extra layers in the FL model to gather more information regarding the parameter server as a malicious attacker. Boenisch et al. \cite{boenisch2023curious} developed an inferential analysis-based attack that relies on the first layer of the target model being fully connected. It can recover the original data in a very short time. However, these methods modify the FL model and are easily detectable. To this end, Zhu et al. \cite{zhu2021rgap} proposed an inferential analysis-based attack that does not change the FL model but suffers from an error accumulation problem, which can degrade the quality of reconstructed data. Therefore, it is necessary to investigate the gradient inversion attack, which not only does not modify the structure and parameters of FL but also breaks through the gradient pruning and adding-noise defense mechanisms.

\section{GI-SMN}\label{sec:method}
This section describes a gradient inversion attack against federated learning using StyleGan-XL \cite{sauer2022stylegan}. It aims to improve attack performance while limiting prior knowledge usage and attacker capabilities.

\subsection{Gradient Inversion Problem}
This work focuses on federated learning for image classification. Eq. \eqref{eq:classLoss} describes a classical supervised classification model, where model $F$ is obtained by minimizing loss function $\ell$ with parameter $w$.
\begin{equation} \label{eq:classLoss}
    \min _{w} \sum_{(x, y) \in D} \ell(F(x, w) , y)
\end{equation}
where $D$ denotes a dataset of ${x}\in \mathbb{R}^{m}$ and label $y\in\left \{ 0,1 \right \} ^{L} $. The corresponding gradient information $\nabla w$ is defined as follows:
\begin{equation} \label{eq:gradient}
    \nabla w=\frac{1}{B} \sum_{i}^{B} \frac{\partial  \ell\left(F\left(x_{i}, w\right), {y}_{i}\right)}{\partial w}
\end{equation}
where $B$ indicates the batch size and $\nabla w$ represents the gradient information obtained by a node using batch images. In FL, all or some nodes send the calculated gradient $\nabla w$ of $\ell(F({x}, w), {y})$ to the parameter server to optimize model $F$.

Inverting the gradient aims to reconstruct the original images utilized to compute the reported gradients. The image in the reconstruction process is referred to as the dummy image. After initialization, the dummy image ${x}^{*}$ is obtained, and the corresponding dummy gradient $\nabla w^{*}$ is generated using Eq. \eqref{eq:gradient}. The image ${x}^{*}$ is reconstructed by the following loss function:
\begin{equation} \label{eq:loss}
    {x}^{*}=\arg \min \left\|\nabla w^{*}-\nabla w\right\|_{F}^{2}+\mathcal{R}_{\mathrm{aux}}({{x}^{*}})
\end{equation}
where $\mathcal{R}_{\mathrm{aux}}$ is the auxiliary regularization, which will be discussed later. When a generative model is introduced in the attack, the result of Eq. \eqref{eq:loss} is used as an input to the generative model to generate a dummy image rather than being directly acquired.

Image reconstruction is comparable to supervised learning, where the true gradient $\nabla w$ corresponds to a high-latitude `label', and the dummy image gradient $\nabla w^{*}$ is the parameter to be learned. Dummy images constantly approach the original images by optimizing the distance between their gradients.

\subsection{The Workflow of GI-SMN}
\begin{figure*}[h]
    \centering
    \includegraphics[width=0.9\textwidth]{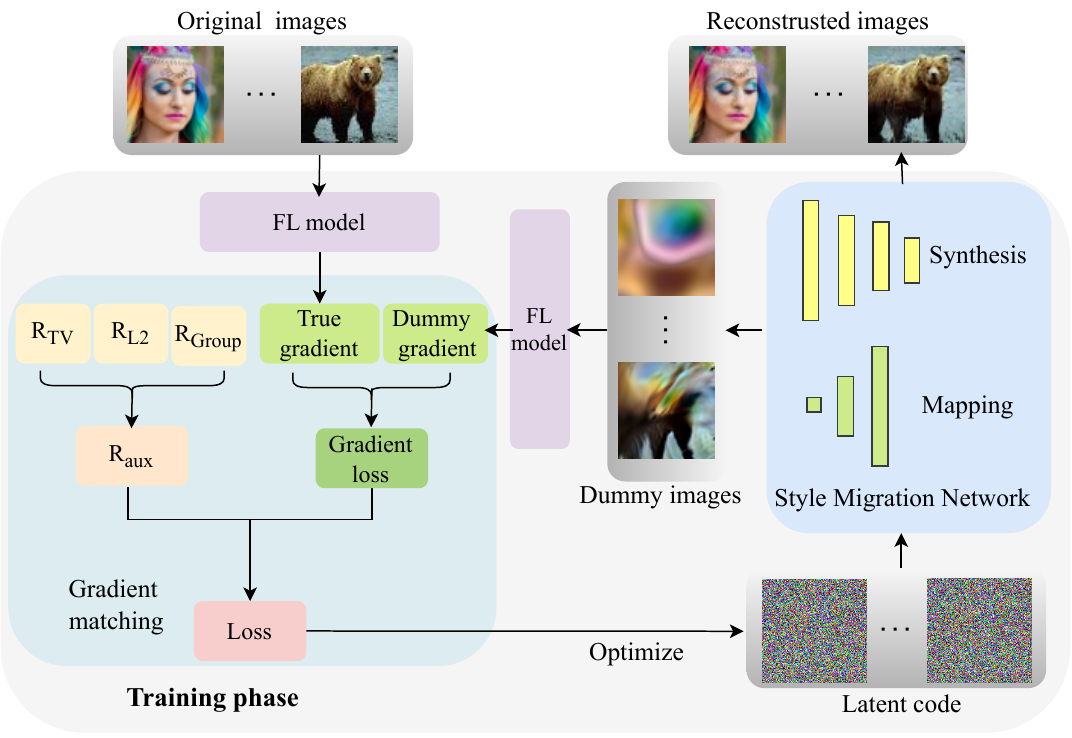}
    \caption{An overview of GI-SMN. GI-SMN optimizes the latent code and employs regularization terms to enhance gradient matching during training. 
    }\label{fig:GI-SMN} \vnudge
\end{figure*}

Fig. \ref{fig:GI-SMN} depicts the main structure of GI-SMN, which comprises image generation, gradient matching, and iterative optimization process. GI-SMN operates in the following steps to successfully reconstruct images from the target FL system. First, a random Gaussian distribution creates latent code fed into the generative model StyleGAN-XL to generate the initial dummy images. Then, the generated dummy images are input into the FL model to get the dummy gradients using backpropagation. Meanwhile, the original images are also fed into the FL model to acquire the true gradients. Next, the latent code is optimized using the loss value given by Eq. \eqref{eq:loss} and placed into the generative model of StyleGan-XL to produce optimized dummy images. When the above iterative optimization reaches a set period, the dummy images are eventually optimized as reconstructed. Algorithm \ref{alg:GI-SMN} provides the details of GI-SMN. 

\begin{algorithm}[htpb] 
	\caption{GI-SMN reconstructs the original image}\label{alg:GI-SMN} 
	\label{alg1} 
	\begin{algorithmic}
		\REQUIRE latent code $z$
		\ENSURE Reconstructed image ${x}^{*}$
		\STATE $z \gets$ {Random initialization}
            \STATE $G \gets$ {Pre-trained StyleGAN-XL model}
            \STATE $\nabla w \gets \frac{1}{B} \sum_{i}^{B} \frac{\partial  \ell\left(F\left(x_{i}, w\right), {y}_{i}\right)}{\partial w} $
		\WHILE{$t < T$}
		\STATE ${x}^{*} \gets G(z)$
		\STATE $\nabla w^{*} \gets \frac{1}{B} \sum_{i}^{B} \frac{\partial  \ell\left(F\left(x^{*}_{i}, w\right), {y}_{i}\right)}{\partial w} $
            \IF{$t < 4T/9$} 
		\STATE $z \gets \arg \min \left\|\nabla w^{*}-\nabla w\right\|_{F}^{2}$
            \ELSE 
            \STATE $\mathcal{R}_{\mathrm{aux}} \gets \alpha_{1} \mathcal{R}_{T V}+\alpha_{2} \mathcal{R}_{L2}+\alpha_{3} \mathcal{R}_{{Group}}$
            \STATE $z \gets \arg \min \left\|\nabla w^{*}-\nabla w\right\|_{F}^{2}+\mathcal{R}_{\mathrm{aux}}$
            \ENDIF
		\ENDWHILE 
		\RETURN ${x}^{*} \gets G(z)$
	\end{algorithmic} 
\end{algorithm}

\subsection{Attacker}
To effectively attack the target with more auxiliary information, traditional gradient inversion attacks tend to assume that the attacker has powerful capabilities \cite{fowl2021robbing,wen2022fishing}. However, the strong assumptions are not realistic in real-world attacks, e.g., collusion with participants can be easily detected \cite{chen2020training}, and tampering with the model or parameters is easily prevented by integrity checks \cite{chen2020training}. In this study, we assume that the attacker is an honest but curious parameter server with access to the gradient information uploaded by each node during FL training. At the same time, the attacker gets access to the FL model and the corresponding parameters to obtain the gradients corresponding to dummy data. This assumption will not interfere with the activities of FL, causing the attack to be more stealthy and thus less noticeable.
Furthermore, previous works demonstrated that the label can be recovered almost precisely by examining the gradient of the last layer \cite{yin2021see}, while the reconstruction of images remains difficult \cite{jeon2021gradient}. We assume the attacker has access to true label information.

\subsection{Trained Generative Model}
There is still room for development for those images reconstructed by existing gradient inversion attacks in terms of quantitative measurements and visual effects \cite{zhang2022survey}. 
To this end, we utilize a specialized pre-trained model, StyleGAN-XL, known for its efficiency in reconstructing high-quality images. This model is particularly advantageous for gradient inversion attacks as it leverages a smaller latent code space compared to the target image's dimensions \cite{jeon2021gradient,sauer2022stylegan}, significantly reducing the required optimization space. For instance, targeting an original image of 3x32x32 pixels, the latent code z in StyleGAN-XL is a constant size of 64 \cite{sauer2022stylegan}, offering a substantial reduction in the optimization domain by a factor of 48. Given the relatively low intrinsic dimensionality of natural image datasets \cite{cao2023fepn}, such as ImageNet, which typically has a dimensionality of approximately 40, the utilisation of a 64-dimensional latent code in StyleGAN-XL simplifies the network's objective of decoupling image features. This, in turn, improves the efficiency and effectiveness of image reconstruction in gradient inversion attacks. Reducing the dimensionality of the optimization target not only significantly decreases computational complexity but also eliminates the need for additional prior knowledge to assist the optimization process. Moreover, it lessens the risk of overfitting. This is because each dimension of the latent code typically controls a broader set of features in images, rather than individual pixels, thus promoting more generalized solutions \cite{sauer2022stylegan}.

\subsection{Auxiliary Regularization}
Regularization terms are utilized to improve the quality of reconstructed images. Here, the regularization terms consist of Total Variation (TV) \cite{geiping2020inverting}, L2 \cite{hatamizadeh2022gradvit}, and Group Regularization \cite{yin2021see}. More formally, the regularization term may be formulated as follows:
\begin{equation}
    \mathcal{R}_{\mathrm{aux}}({x})=\alpha_{1} \mathcal{R}_{T V}({x})+\alpha_{2} \mathcal{R}_{L2}({x})+\alpha_{3} \mathcal{R}_{{Group }}({x})
\end{equation}
where the scaling factors $\alpha_{1}$, $\alpha_{2}$, and $\alpha_{3}$ balance the regularization terms. 

TV regularization is an image processing regularization approach that achieves image smoothing by regulating the total variation of images. In gradient inversion, TV regularization may minimize noise and recover detailed information about images by limiting the difference between pixels and suppressing high-frequency noise components in the images \cite{geiping2020inverting,jeon2021gradient}. More formally, TV regularization can be expressed as Eq. \eqref{eq:regular}.
\begin{equation}\label{eq:regular}
    \mathcal{R}_{\mathrm{TV}}({x})=\sum_{(i, j)} \sum_{\left(i^{\prime}, j^{\prime}\right) \in \partial(i, j)}\left\|{x}(i, j)-{x}\left(i^{\prime}, j^{\prime}\right)\right\|^{2}
\end{equation}
where $\partial(i, j)$ denotes a set of neighboring pixels of $(i, j)$ in images. 

Group regularization also be utilized to improve the quality of reconstructed images. Gradient inversion attacks are brutal to do with convolution neural networks (CNNs) because dummy images have to be precisely placed about the target images. Convolution, pooling, and filling operations cause spatial variance in the forward pass of CNNs, making it challenging to locate them in image reconstruction. Group regularization can alleviate this problem \cite{hatamizadeh2022gradvit}, as described below:
\begin{equation}
    \mathcal{R}_{\text {Group }}({{x}_{i}})=\left\|{{x}_{i}}-\mathbb{E}\left({{x}}_{g \in G}\right)\right\|^{2}
\end{equation}
where ${x}_{i}$ is a single image in a batch. $\mathbb{E}({{x}}_{g \in G})$ computes the average of the pixels of the entire batch of images generated by the generative model $G$. It penalizes any candidate $x_i$ once it deviates away from the group's consensus images $\mathbb{E}({{x}}_{g \in G})$. To avoid calculating $\mathbb{E}({{x}}_{g \in G})$ frequently, we update $\mathbb{E}({{x}}_{g \in G})$ every 100 epochs during optimization. 

Unlike the Group term, the L2 regular term constrains the variation range of a single image \cite{hatamizadeh2022gradvit}, stated as Eq. \eqref{eq:L2}. The localization problem is alleviated by combining L2 with the Group regular term to achieve better reconstruction results.
\begin{equation} \label{eq:L2}
    \mathcal{R}_{\text {L}_{2}}({x})=\left\|{{x}_{i}}\right\|^{2}
\end{equation}

The loss function is crucial for reconstructing original images. In the early stages of iterations, the regularization auxiliary term with gradient loss leads to convergence to a sub-optimal solution \cite{hatamizadeh2022gradvit}. Therefore, the action time of regularization terms is adjusted to reduce the probability of falling into a sub-optimal solution. For a total of $T$ training iterations, the loss scheduler at iteration $t$ can be defined as Eq. \eqref{eq:interation}.
\begin{equation}\label{eq:interation}
\Gamma(t)=\left\{
    \begin{array}{ll}
    \alpha_{\text {grad }} & 0<t<\frac{4}{9} T \\
    \frac{1}{2} \alpha_{\text {grad }}+\alpha_{\text {aux }}  & \frac{4}{9} T<t<T
    \end{array}\right.
\end{equation}
where $\alpha_{\text {grad }}$ and $\alpha_{\text {aux }}$ signify the gradient loss function and $\mathcal{R}_{\mathrm{aux}}$ matching scale factors, respectively. After numerous experiments, we set experience values of $1/2$ and $4/9$ for them.

\section{Performance Evaluation}\label{sec:experiment}
\subsection{Experimental Setup}

\subsubsection{Datasets description.}
To assess the effectiveness and breadth of GI-SMN, we conduct image classification tasks on two widely used datasets, CIFAR10 and ImageNet, while performing face recognition tasks on the famous face datasets, FFHQ. Due to the massive number of parameters in the pre-trained model of StyleGAN-XL, we scale down images to 32x32 pixels to reduce the computational overhead. Furthermore, we utilize TinyImageNet \cite{huang2023neurogenesis}, a subset of ImageNet, to minimize computational effort even more. 

\subsubsection{Evaluation metrics.}
To objectively measure the quality of reconstructed images, we utilize PSNR \cite{zhang2018unreasonable} as a quantitative analysis metric. PSNR is a widely used metric in gradient inversion that assesses the similarity between the reconstructed and original images. It is formulated as Eq. \eqref{eq:psnr}.
\begin{equation}
\label{eq:psnr}
PSNR=20\cdot\log_{10}\left(\frac{MAX_{I}}{\sqrt{MSE}}\right)
\end{equation}
where $\rm MAX_{I}$ denotes the maximum pixel value of images, and MSE represents the mean square error. Typically, with a PSNR value over 30, the image quality is excellent, and the dummy image is very similar to the original image.

We also adopt LPIPS and SSIM \cite{zhang2018unreasonable} to evaluate the visual similarity of two images based on perceptual information. LPIPS evaluates the visual similarity of two images based on perceptual information. In contrast to PSNR, a smaller LPIPS value indicates a higher similarity. In addition, SSIM also measures the similarity of two images. Its value ranges from -1 to 1, with a value close to 1 indicating high image similarity.

\subsubsection{Implementation details.}
We use a randomly initialized ResNet18 \cite{jeon2021gradient} as the target model of FL. The StyleGAN-XL is a pre-trained model trained on ImageNet and consists of two sub-networks: Mapping and Synthesis. We set the default batch size to 4 and utilized the MSE as a loss function to measure the difference between gradients. Additionally, we apply the Adam optimizer with a learning rate of 0.1 for 4000 epochs, adjusting the learning rate at 3/8, 5/8, and 7/8 of Epochs. The balance factor for all three regularization terms is set to 1E-4. We initialize the latent code using the standard normal distribution and place the seed at 1314. The above hyperparameter values are empirical values from experiments. We employ two NVIDIA 3090 for computing in all experiments.

\subsubsection{Comparison gradient inversion attacks.}
We compare GI-SMN with the following state-of-the-art gradient inversion attacks. 
\begin{enumerate}[label=(\arabic*)]
    \item Inverting gradients (IG) \cite{geiping2020inverting}. It takes the cosine distance as loss and utilizes the total variation as regular terms. It also makes use of the Adam for optimization.
    
    \item Fishing \cite{wen2022fishing}. It employs gradient amplification to find target data in large-batch federated learning.
    
    \item GIAS \cite{jeon2021gradient}. Like IG, it incorporates the BN statistics model and StyleGAN2 model to optimize the reconstruction of dummy images.
\end{enumerate}

We implemented these attacks according to the code repositories and parameters released by the authors \cite{geiping2020inverting,jeon2021gradient,wen2022fishing}. For a fairer comparison, we normalize the FL model to ResNet18 and provide true labels \cite{jeon2021gradient}. Additionally, we do not impose any restrictions on the comparative methods regarding the use of image prior knowledge and the proactive capabilities of the attackers.

\begin{table*}[t]
\normalsize
    \caption{A comparison of experimental results generated by GI-SMN versus state-of-the-art gradient inversion attacks on three datasets and batches. Since Fishing can only reconstruct a single image, only one PSNR value is shown in the table and the mean value of the other methods is demonstrated under batch.}  
    \label{tab1}
    \centering
    \setlength{\tabcolsep}{0.1cm} 
    \renewcommand{\arraystretch}{1.25} 
    \resizebox{\textwidth}{!}{
    \begin{tabular}{ccccccccccc}
        \toprule
    & \multirow{2}{*}{{Attacks}}   &   & {CIFAR10} &    &    & {FFHQ} &    &    &{ImageNet}  &  \\
    &    & {PSNR$\uparrow$} & {SSIM$\uparrow$}    & {LPIPS$\downarrow$} & { PSNR$\uparrow$} & {SSIM$\uparrow$} & {LPIPS$\downarrow$} & {PSNR$\uparrow$} & {SSIM$\uparrow$}    & {LPIPS$\downarrow$} \\
    \midrule
    \multirow{4}{*}{{Batch=1}} 
    & {IG} & 24.47   & 0.91   &  0.0024    & 23.36   &  0.92  &  0.0055  &22.83 &0.92 &0.0054   \\
    & {Fishing}  & 10.11  & 0.43  &0.1100  &  14.66  & 0.48  &0.1000  &16.08  &0.57 &0.0700    \\
    & {GIAS}  & 32.79   &  0.98   & 0.0006   & \textbf{34.46}    & \textbf{0.99}  &\textbf{0.0003} &32.27  &0.98  &0.0013 \\
    &{GI-SMN}   &\textbf{36.98} & \textbf{0.99}   & \textbf{0.0003}   & 31.31   & \textbf{0.99}   & \textbf{0.0003} &\textbf{34.30}  &\textbf{0.99}  &\textbf{0.0004}    \\
    \midrule
    \multirow{4}{*}{{Batch=4}} 
    & {IG} & 17.35   & 0.67  & 0.0299  & 18.45  & 0.65   & 0.0400 &17.32   &0.64  &0.0400    \\
    & {Fishing}   & 14.82   & 0.39    & 0.5000   & 13.06    & 0.33    &0.4600  &13.53  &0.25  &0.4500   \\
    & {GIAS}     & 25.02    & 0.91    & 0.0046   & 25.34    & 0.94    &0.0039  &28.36  &0.95  &0.0042   \\
    & {GI-SMN}    & \textbf{29.40}      & \textbf{0.96}    & \textbf{0.0033}     &\textbf{29.07}    &\textbf{0.96}    &\textbf{0.0017}  &\textbf{31.84}  &\textbf{0.97}  &\textbf{0.0018}    \\
    \midrule
    \multirow{4}{*}{{Batch=6}} 
    & {IG} & 15.41    & 0.56    & 0.0543    & 15.67    & 0.57   & 0.0620   &15.64  &0.57  &0.0757  \\
    & {Fishing}    &12.56     & 0.41     & 0.5400    & 12.59     &0.38    & 0.5000 &10.51  &0.53  &0.5600   \\
    & {GIAS} & 21.06    & 0.82   &0.0130    &25.02   &0.92   & 0.0065 &24.91  &0.89  &0.0123  \\
    & {GI-SMN}  &\textbf{28.01}   &\textbf{0.95}   & \textbf{0.0041}   & \textbf{28.93}  &\textbf{0.96}    &\textbf{0.0016}  &\textbf{29.45}  &\textbf{0.96}  & \textbf{0.0100}   \\
        \bottomrule
\end{tabular}}
\end{table*}

\subsection{Results Analysis}\label{sec:result}

\subsubsection{Comparison with state-of-the-art gradient inversion attacks.}

We performed gradient inversion attack comparison experiments on three datasets to reconstruct images with different data batches and record the experimental results as shown in Table \ref{tab1}. In the three different datasets, our method GI-SMN obtains the best results for different batches of data reconstruction, with slightly better PSNR values for the GIAS method outperforming ours for a batch of 1 on the FFHQ dataset only. In contrast, the same values are obtained for the LPIPS and SSIM metrics. Compared to IG and Fishing, GI-SMN achieves an average of 125\% and 61\% PSNR value improvement. For the GIAS method, whose results are close to our process, we achieved an average of 10\% PSNR value improvement on different batches of the three datasets, with a maximum of 33\% improvement on individual data. Overall, GI-SMN achieves the highest similarity, which means that our reconstructed images possess superior fidelity. 

The comparison experiments also reveal the effect of data variability on the reconstruction results. For example, in Table \ref{tab1}, our method GI-SMN typically performs better on the TinyImageNet images than on the FFHQ dataset, especially for small batch reconstruction. This discrepancy primarily arises because FFHQ face images contain a higher level of detail compared to the less complex TinyImageNet images, presenting additional challenges for gradient inversion. However, the effect of this difference gradually decreases when the batch size increases, mainly because the difficulty caused by the increase outweighs the impact of the difference in the data itself.

\begin{figure}[htb!]
    \centering
   \includegraphics[width=0.6\columnwidth]{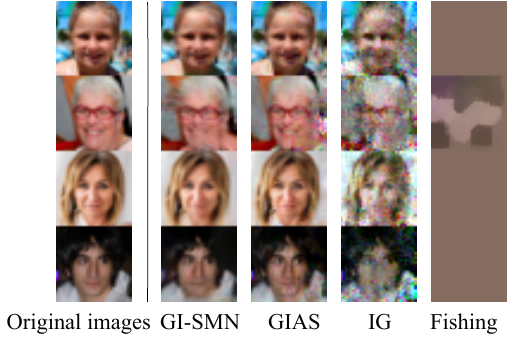}
    \caption{A comparison of images reconstructed by GI-SMN and state-of-the-art gradient inversion attacks with a batch size of four on FFHQ. The left column represents images from FFHQ, and the rest denote images constructed by different gradient inversion attacks. 
    }\label{fig:vision}\vnudge
\end{figure}

Furthermore, Fig. \ref{fig:vision} reveals the reconstructed face images from different gradient inversion attacks on FFHQ. While the GIAS achieves a complete reconstruction, its results still contain noise and distortion. The IG method generates significantly noisy images, which seriously affects the recognizability of its images. Fishing cannot reconstruct batch images due to its strategy of modifying parameters. In contrast, the GI-SMN method excels in preserving details and maximizes restoring the original image's features. Even under more stringent constraints imposed on the attacker and the required prior knowledge, GI-SMN outperforms the other compared methods regarding data fidelity and visualization. The above results further validate the superiority of our process in image reconstruction similarity metrics obtained.

\begin{figure*}[htp!]
    \centering  
    \subfloat[Similarity comparison]{\label{fig:ImageSizeA}
        \includegraphics[width=0.44\textwidth]{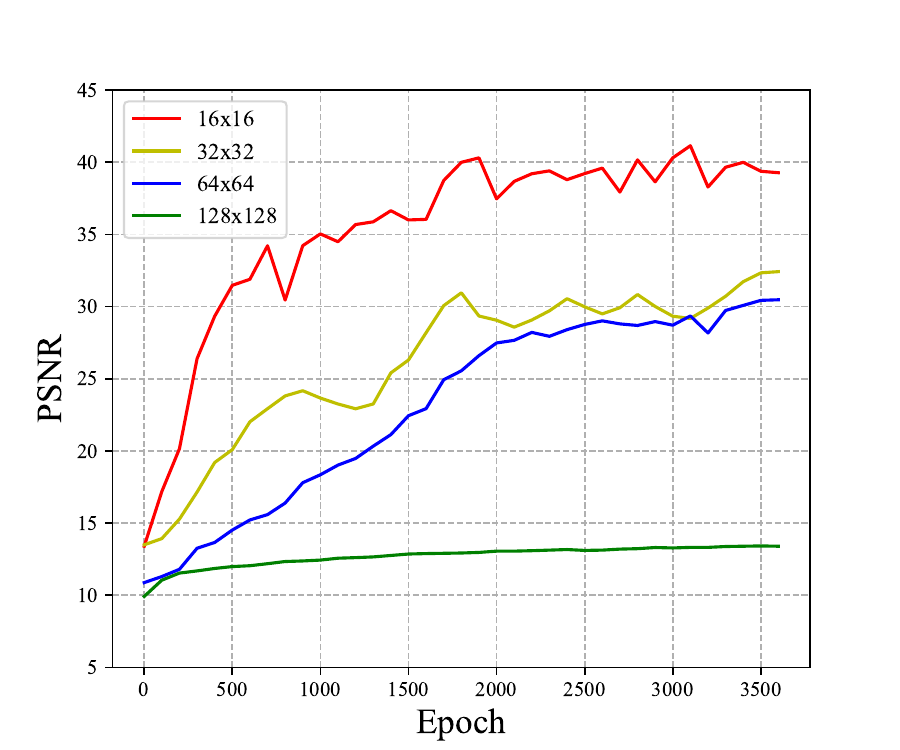}}
    \subfloat[Visual comparison]{\label{fig:ImageSizeB}
       \includegraphics[width=0.48\textwidth]{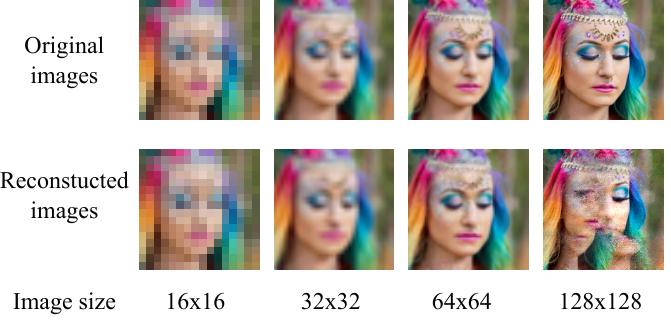}}
    \caption{Effects of reconstructing images of different sizes in PSNR similarity and visual comparison. 
    }
    \label{fig:ImageSize}\vnudge
\end{figure*}

\subsubsection{Effect of image size on image reconstruction.}
The pixel size of an image is a crucial factor affecting the reconstruction results of gradient inversion. We evaluate the reconstruction effect of the GI-SMN method on images with different resolutions, where Fig. \ref{fig:ImageSizeA} demonstrates the variation of PSNR values during the reconstruction process, while Fig. \ref{fig:ImageSizeB} presents the final reconstructed image. Fig. \ref{fig:ImageSizeA} reveals an intuitive trend: as the image resolution increases, the PSNR values tend to decrease, confirming that higher pixel images pose more significant challenges in the gradient inversion reconstruction process. Notably, for images with resolutions up to 64x64 pixels, the PSNR values of the reconstructed images generally exceed 30, as shown in Fig. \ref{fig:ImageSizeB}, indicating these images successfully restore the details and colors of the original images. When dealing with 128x128 pixel resolution images, although some confusion appears in the reconstructed images, the key visual features and colors are still relatively accurately captured and reproduced. Our method demonstrates effective reconstruction capabilities for 64x64 pixel images and shows potential for reconstructing 128x128 pixel images.

\subsubsection{Effect of batch size and loss function on image reconstruction.}
\begin{figure*}[t]
    \centering  
    \subfloat[Batch size]{\label{fig:BatchSize}
        \includegraphics[width=0.49\textwidth]{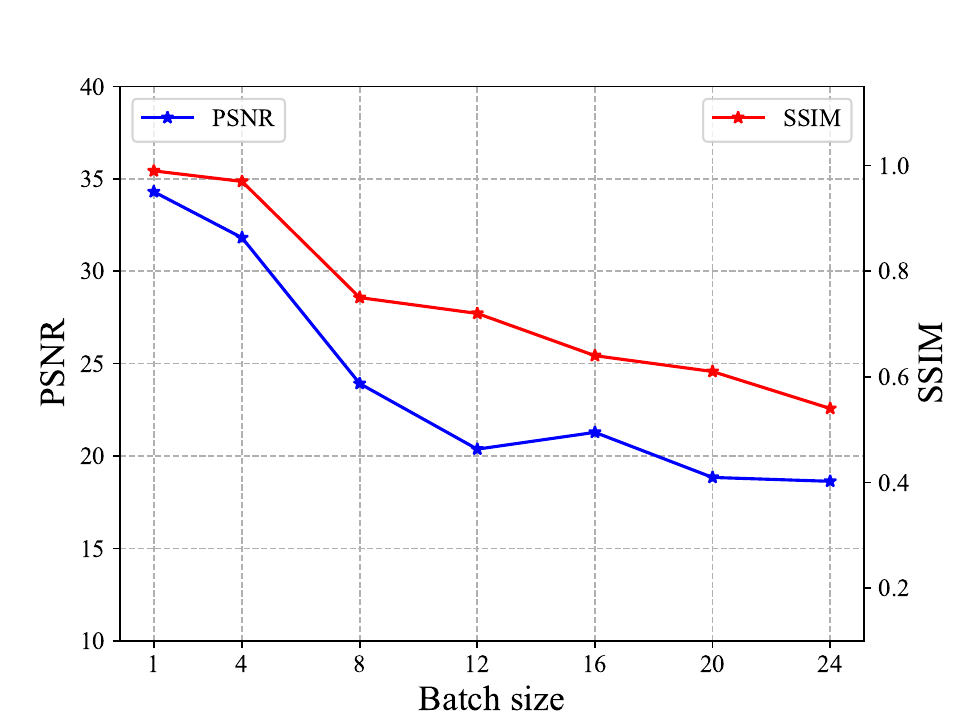}} 
    \subfloat[Loss functions]{\label{fig:loss}
        \includegraphics[width=0.44\textwidth]{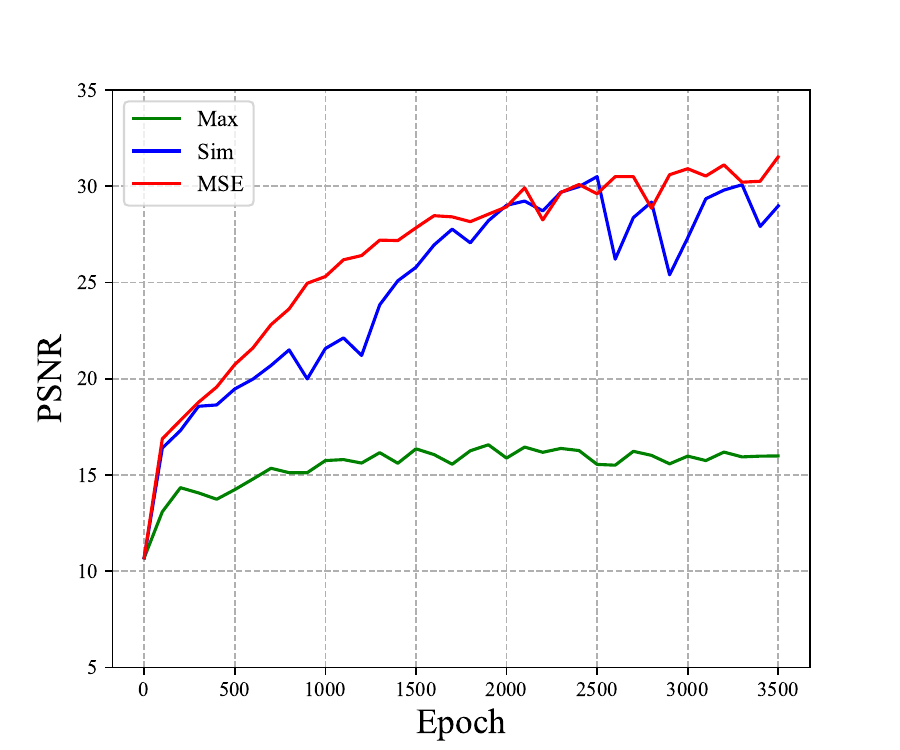}}
    \caption{The effect of different batch sizes and loss functions on gradient inversion. GI-SMN utilizes an optimal loss function that enables efficient batch reconstruction of images.}
    \label{fig:BatchSize and loss functions}\vnudge
\end{figure*}

The batch size of the reconstructed images is also a key factor affecting gradient inversion. Table \ref{tab1} shows some batch comparisons where GI-SMN achieves leading results at different batches. We further extend this with more batch size experiments in Fig. \ref{fig:BatchSize}. From these experiments, it is observed that the PSNR and SSIM values show a similar decreasing trend as the batch size of the reconstructed image increases, indicating that the gradient inversion attack becomes more and more difficult with the increase in batch size, which inspires the use of larger batch values as a safer strategy in federated learning. It is also worth noting that when the batch size does not exceed 16, our reconstructed images have PSNR values of more than 20 and SSIM values of more than 0.6, which show a more effective reconstruction for small batch samples. However, limited by the high memory requirement of the generative network, we could not realize further experiments on larger batches of samples. This limitation reveals one of the technical challenges that need to be considered when conducting experiments on gradient inversion attacks and also points to issues that need to be addressed in future research.

We further analyze the effect of different loss functions on the effectiveness of gradient inversion attacks during gradient matching. According to the experimental results shown in Fig. \ref{fig:loss}, the cosine similarity (Sim) and the MSE perform similarly in providing an upper bound on the reconstruction effect. At the same time, the MSE has performed better in terms of stability. Comparatively, the Maximum Absolute Error (Max) is significantly less effective than the other methods. This phenomenon is because the Max loss function only considers the maximum absolute value difference in the gradient and ignores other gradient information, resulting in a degradation of the reconstruction quality.

\subsubsection{Effectiveness of GI-SMN against gradient pruning and gradient interception.}

\begin{figure*}[tp!]
    \centering  
    \subfloat[Gradient pruning]{\label{fig:pruning}
        \includegraphics[width=0.48\textwidth]{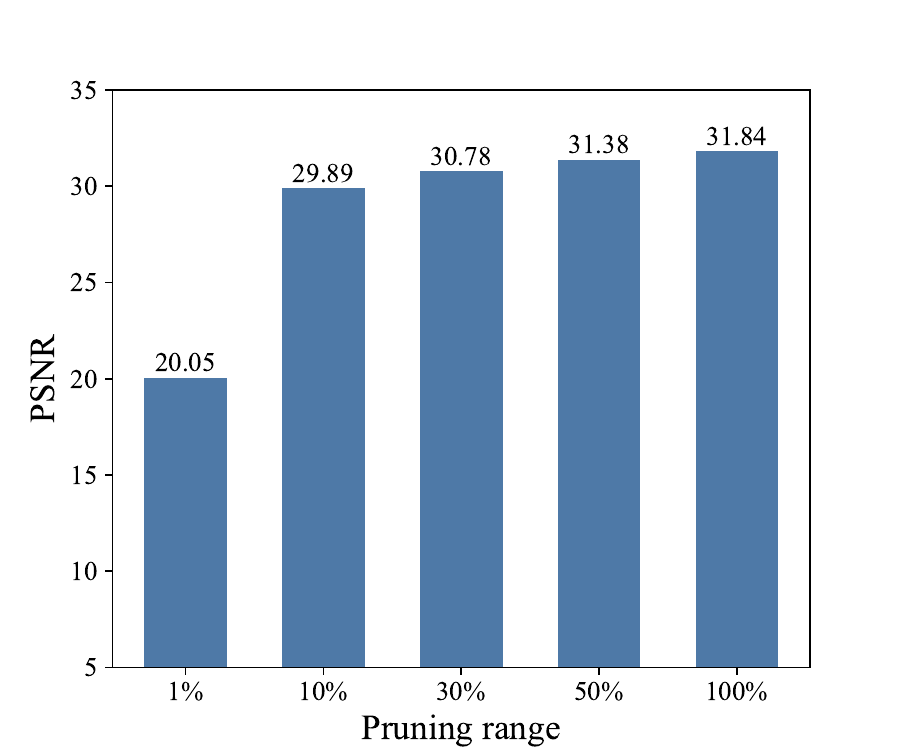}}
    \subfloat[Gradient interception]{\label{fig:interception}
        \includegraphics[width=0.48\textwidth]{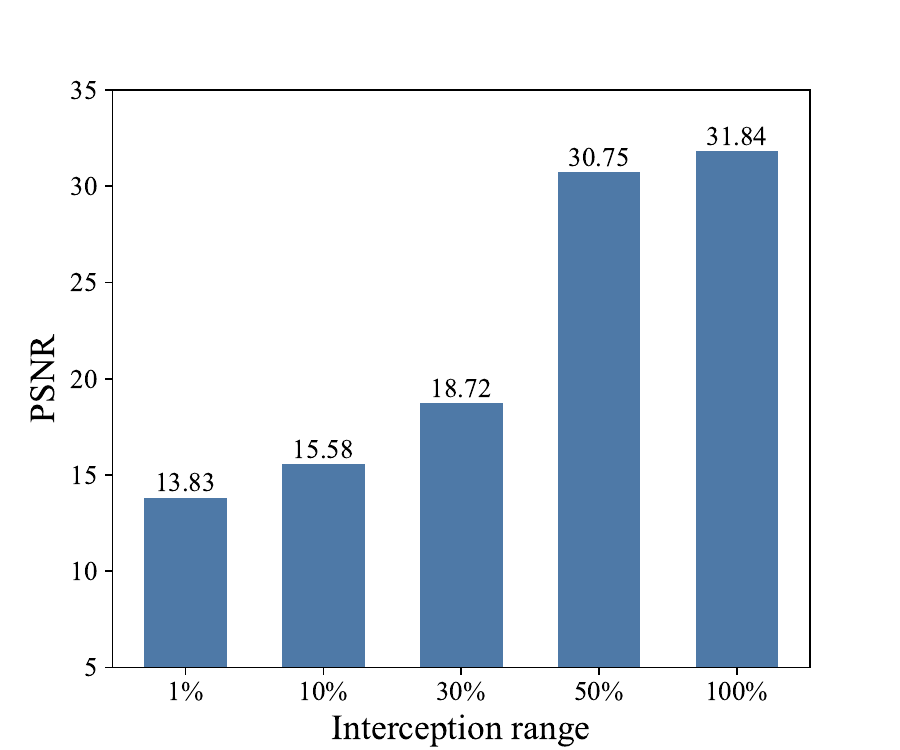}}
    \caption{The impact of gradient pruning and gradient interception on GI-SMN. The horizontal axis indicates the retained gradient information (\%). 
 }\vnudge
\end{figure*}

Since FL relies on the exchange of gradient information, a straightforward FL privacy-preserving approach is to protect gradients. Gradient pruning protects against gradient inversion attacks by setting smaller gradients to zero. Zhu et al. \cite{zhu2019deep} demonstrated that pruning more than 70\% of gradients can cause gradient inversion failure. We evaluate the effectiveness of our method under the gradient pruning defense approach as shown in Fig. \ref{fig:pruning}, where the horizontal coordinate indicates the percentage of gradient information retained. We found that the score of the reconstructed image can still reach 20.05 even when only the maximum 1\% of gradient information is used. Gradient pruning at the 90\% level caused only a 6\% loss of score, and when the pruning rate is below 70\%, it hardly affects the reconstruction. Our attack method can still achieve effective reconstruction under gradient pruning defense. This finding also reveals that the more considerable absolute value of gradient information has a decisive role in the reconstruction effect in the gradient inversion process.

We also evaluated another form of gradient truncation, such as intercepting 10\% of the gradient vector and setting the remaining gradient information to 0, distinguishing it from gradient pruning by selecting the percentage gradient information of the maximum value. As shown in the right panel of Fig. \ref{fig:interception}, this type of truncation has a significant negative impact on the effectiveness of the attack, especially when the first 30\% of the gradient is retained, which can significantly reduce the effectiveness of the attack. However, even so, when keeping the first 50\% of the gradient information, our attack achieves similar effectiveness as when there is no defense. These results suggest that while gradient pruning and truncation can reduce the effectiveness of gradient inversion attacks to some extent, they are not foolproof solutions against efficient attack strategies. This emphasizes developing more advanced defense strategies to ensure data privacy security in federated learning environments.

\begin{figure}[h]
    \centering
   \includegraphics[width=0.5\textwidth]{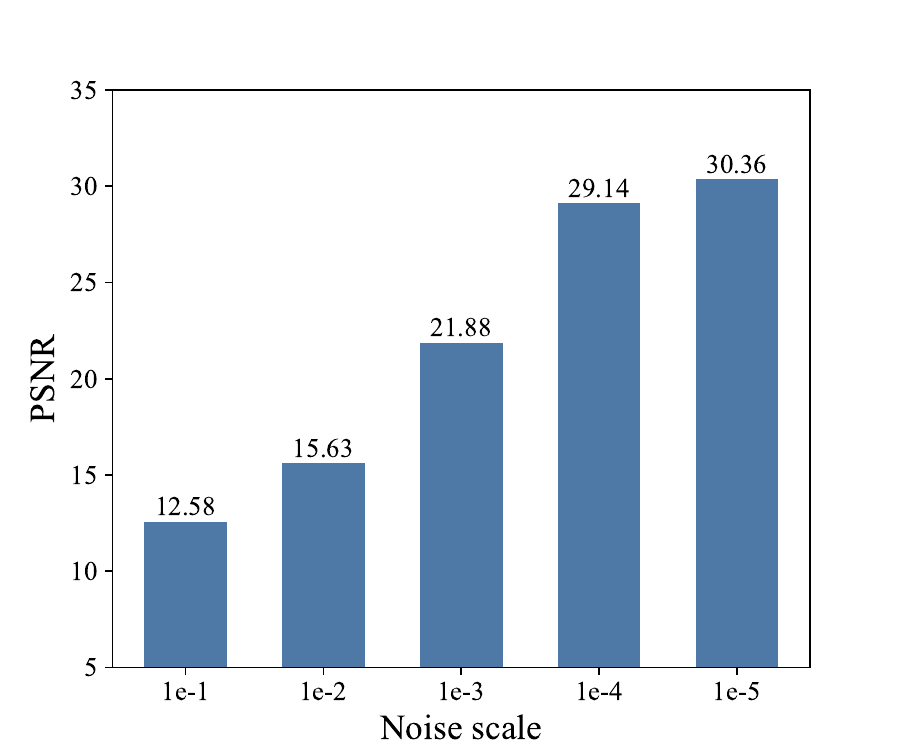}
    \caption{The effect of GI-SMN against the differential privacy defense. The horizontal axis indicates the level of noise added to the gradient information.
    }\label{fig:noise} \vnudge
\end{figure}

\subsubsection{Effectiveness of GI-SMN against differential privacy.}

Incorporating noise into the gradient sharing process is a widely used differential privacy defense to enhance against gradient inversion attacks \cite{abadi2016deep,zhu2019deep}. To evaluate the performance of the gradient inversion attack under different noise levels, Gaussian noise with 0 as the mean and variance varying from 1e-1 to 1e-5 is introduced in our experiments. As can be seen in Fig. \ref{fig:noise}, the reconstruction results reduce the metrics as the noise increases, reflecting that the effectiveness of the gradient noise defense depends on the magnitude of the noise variance. It is worth noting. When the noise variance is less than 1e-4, the effect on the gradient reconstruction effect is minimal. When the noise variance is 1e-3, it achieves very good results, with a PSNR value of 21.88. In contrast, an increase in the variance to more than 1e-2 usually seriously affects the regular optimization of the model, and it is difficult for gradient inversion to reconstruct the original image effectively. In summary, the above results show that the gradient inversion attack is still a threat even in the presence of noise interference. Further research and robust defense measures are needed to ensure data privacy security.

\begin{table}[ht]
\normalsize
    \setlength{\tabcolsep}{7pt} 
    \caption{Effect of different initialization functions and random seeds on gradient inversion. Randn denotes the standard normal distribution, Rand represents the uniform distribution, and Normaly is the normal distribution constrained by the ImageNet mean and variance.}  
    \label{tab:las}
    \centering
    \begin{tabular*}{0.6\columnwidth}{ccccc}
    \toprule
    Function  &  Seed & GI-SMN &  GIAS  &  IG    \\
    \midrule
    \multirow{3}{*}{{Randn}} 
    & {88}    & 31.63   & 28.24   & 19.90    \\
    & {1314}  & 31.84  & 28.36   & 16.84    \\
    & {2586}  & 31.98   & 30.19    &16.87   \\
    \midrule
    \multirow{3}{*}{{Rand}} 
    & {88}    & 30.51   & 28.44   &  18.30     \\
    & {1314}  & 31.57  & 29.21  &  17.43   \\
    & {2586}  & 30.78  & 29.79    &  17.75 \\
    \midrule
   \multirow{3}{*}{{Normal}} 
    & {88}    & 30.91  & 29.66   &  21.67     \\
    & {1314}  & 31.18  & 28.67   & 20.24    \\
    & {2586}  & 30.59  & 29.42    & 16.73  \\
    \bottomrule
\end{tabular*}
\end{table}

\subsubsection{Effectiveness of dummy image initialization.}
In gradient inversion attacks, the initialization of the virtual image has a significant impact on the final reconstruction results, especially when random initialization is used. The reconstruction often appears unstable, and in some cases, the rebuilding fails \cite{yin2021see}. Therefore, we tested the stability of the GI-SMN method in different initialization methods, as shown in Table \ref{tab:las}. The data in Table 3.4 show that the fluctuations in the PSNR values of our GI-SMN reconstruction samples are stable within 5\% at different random initialization methods. In contrast, the fluctuations are close to 7\% for GIAS and close to 30\% for IG. This result demonstrates our method's excellent stability under different initialization scenarios, mainly due to our framework of relying on generative network models and appropriately designed regularization terms. Together, these mechanisms effectively mitigate the effects of random initialization and ensure that the attack method obtains consistent and reliable reconstruction results even under different initialization conditions. The GIAS method exhibits good stability based on the generative network model. However, the fact that GI-SMN differs in its regularization strategy and network structure makes it more robust in dealing with initialization changes, thus outperforming GIAS in terms of stability. This stability advantage is significant for gradient inversion attacks as it ensures that the attacker obtains reliable reconstruction results even under uncertain initialization conditions, thus increasing the attack's practicality and effectiveness.

\section{Conclusions}
\label{sec:conclusion}
Gradient inversion attacks have recently attracted widespread attention since they pose a significant privacy threat to federated learning. In this paper, we propose an innovative gradient inversion attack based on a style migration network against FL, which requires neither super-attackers nor idealized prior knowledge as traditional gradient attacks. In GI-SMN, we reduce the optimization space by optimizing the latent code and enhance the image reconstruction capability by using regularization terms. Extensive experiments have been conducted on three different datasets. Experimental results demonstrate that GI-SMN outperforms existing state-of-the-art gradient inversion attack methods in visualization and similarity metrics.
Moreover, we also prove that gradient pruning and differential privacy are not effective defences against privacy breaches in federated learning.

In this paper, we conduct a preliminary exploration of gradient inversion attacks against federated learning, which do not require super attackers and additional prior knowledge. The observed attack outcomes are favourable. This illustrates trends that can be utilised to direct future efforts towards more comprehensive privacy-preserving federated learning methodologies. Our forthcoming endeavours will adhere to this path, as we persist in our pursuit of novel systems that may use advancements in federated learning while mitigating the introduction of vulnerabilities that may result in gradient inversion attacks.

\bibliographystyle{splncs04}
\bibliography{ref}

\end{document}